\documentclass[conference]{IEEEtran}

\usepackage{cite}
\usepackage{amsmath,amssymb,amsfonts}
\usepackage{graphicx}
\usepackage{textcomp}
\usepackage{xcolor}

\newtheorem{definition}{Definition}[section] % reset theorem numbering for each chapter
\usepackage{booktabs} % For formal tables
\usepackage{color}
\usepackage{multirow}
\usepackage{subfigure}
\usepackage[ruled]{algorithm2e}

\usepackage[T1]{fontenc}
\usepackage{fix-cm}

\def\BibTeX{{\rm B\kern-.05em{\sc i\kern-.025em b}\kern-.08em
    T\kern-.1667em\lower.7ex\hbox{E}\kern-.125emX}}
\begin{document}

% \title{Paper Title*\\

%\title{Disaster Victim Localization and Rescue Coordination via Reinforcement Learning}
%\title{Disaster Rescuing via Reinforcement Learning}
%\title{Improving the Coordination of Disaster Emergency Response Efforts with Heuristic Reinforcement Learning}
\title{Coordinating Disaster Emergency Response with Heuristic Reinforcement Learning}

% single blind review, so authors should be hidden$
\author{
\IEEEauthorblockN
{
    Long H. Nguyen,\IEEEauthorrefmark{1}
    Zhou Yang,\IEEEauthorrefmark{1}
    Jiazhen Zhu,\IEEEauthorrefmark{2}
    Jia Li,\IEEEauthorrefmark{3}
    Fang Jin\IEEEauthorrefmark{1}
}

\IEEEauthorblockA{\IEEEauthorrefmark{1} Department of Computer Science, Texas Tech University}
\IEEEauthorblockA{\IEEEauthorrefmark{2} George Washington University}
%, Lubbock, Texas, USA
\IEEEauthorblockA{\IEEEauthorrefmark{3} Department of Civil, Environmental and Construction Engineering, Texas Tech University}
    \{
        long.nguyen,
        zhou.yang,
        jia.li,
        fang.jin
    \}@ttu.edu,
    \{
        jiazhen\_zhu
    \}@gwu.edu
}

% \author{
%     \IEEEauthorblockN{
%         Author One\IEEEauthorrefmark{1},
%         Author Two\IEEEauthorrefmark{2}, 
%         Author Three\IEEEauthorrefmark{3} and
%         Author Four\IEEEauthorrefmark{4}}
%     \IEEEauthorblockA{Department of Whatever,
% Whichever University\\
% Wherever\\
% Email: 
%     \IEEEauthorrefmark{1}author.one@add.on.net,
%     \IEEEauthorrefmark{2}author.two@add.on.net,
%     \IEEEauthorrefmark{3}author.three@add.on.net,
%     \IEEEauthorrefmark{4}author.four@add.on.net}
% }

% \author{\IEEEauthorblockN{1\textsuperscript{st} Yang, Nguyen}
% \IEEEauthorblockA{\textit{Department of Computer Science} \\
% \textit{Texas Tech University}\\
% Lubbock, Teas, USA \\
% {yang.zhou, long.nguyen}@ttu.edu}
% \and
% \IEEEauthorblockN{2\textsuperscript{nd} Zhu}
% \and
% \IEEEauthorblockN{3\textsuperscript{rd} Li}
% \IEEEauthorblockA{
% \textit{TechMRT Center}\\
% \textit{Department of Civil, Environmental}\\
% \textit{and Construction Engineering}\\
% \textit{Texas Tech University}\\
% Lubbock, Texas, USA \\
% jia.li@ttu.edu}
% \and
% \IEEEauthorblockN{4\textsuperscript{th}Jin}
% \IEEEauthorblockA{\textit{Computer Science Department} \\
% \textit{Texas Tech University}\\
% Lubbock, Texas \\
% fang.jin@ttu.edu}
% }

\newcommand{\theName}{ResQ}

\maketitle
\IEEEpeerreviewmaketitle

\begin{abstract}
A crucial and time-sensitive task when any disaster occurs is to rescue victims and distribute resources to the right groups and locations. This task is challenging in populated urban areas, due to the huge burst of help requests generated in a very short period. To improve the efficiency of the emergency response in the immediate aftermath of a disaster, we propose a heuristic multi-agent reinforcement learning scheduling algorithm, named as \theName{}, which can effectively schedule the rapid deployment of volunteers to rescue victims in dynamic settings.
The core concept is to quickly identify victims and volunteers from social network data and then schedule rescue parties with an adaptive learning algorithm. This framework performs two key functions: 1) identify trapped victims and rescue volunteers, and 2) optimize the volunteers' rescue strategy in a complex time-sensitive environment. 
The proposed \theName{} algorithm can speed up the training processes through a heuristic function which reduces the state-action space by identifying the set of particular actions over others.
Experimental results showed that the proposed heuristic multi-agent reinforcement learning based scheduling outperforms several state-of-art methods, in terms of both reward rate and response times.

%we proposed a disaster relief framework using social media data and reinforcement learning, which is tested on data collected during the 2017 Hurricane Harvey. 
% an efficient Reinforcement learning framework by mining social media data. 
\end{abstract}

% \begin{IEEEkeywords}
% Disaster Relief, Heuristic Reinforcement Learning, Resource Management
% \end{IEEEkeywords}

\section{Introduction}
\begin{figure}
\centering
\includegraphics[width=\linewidth]{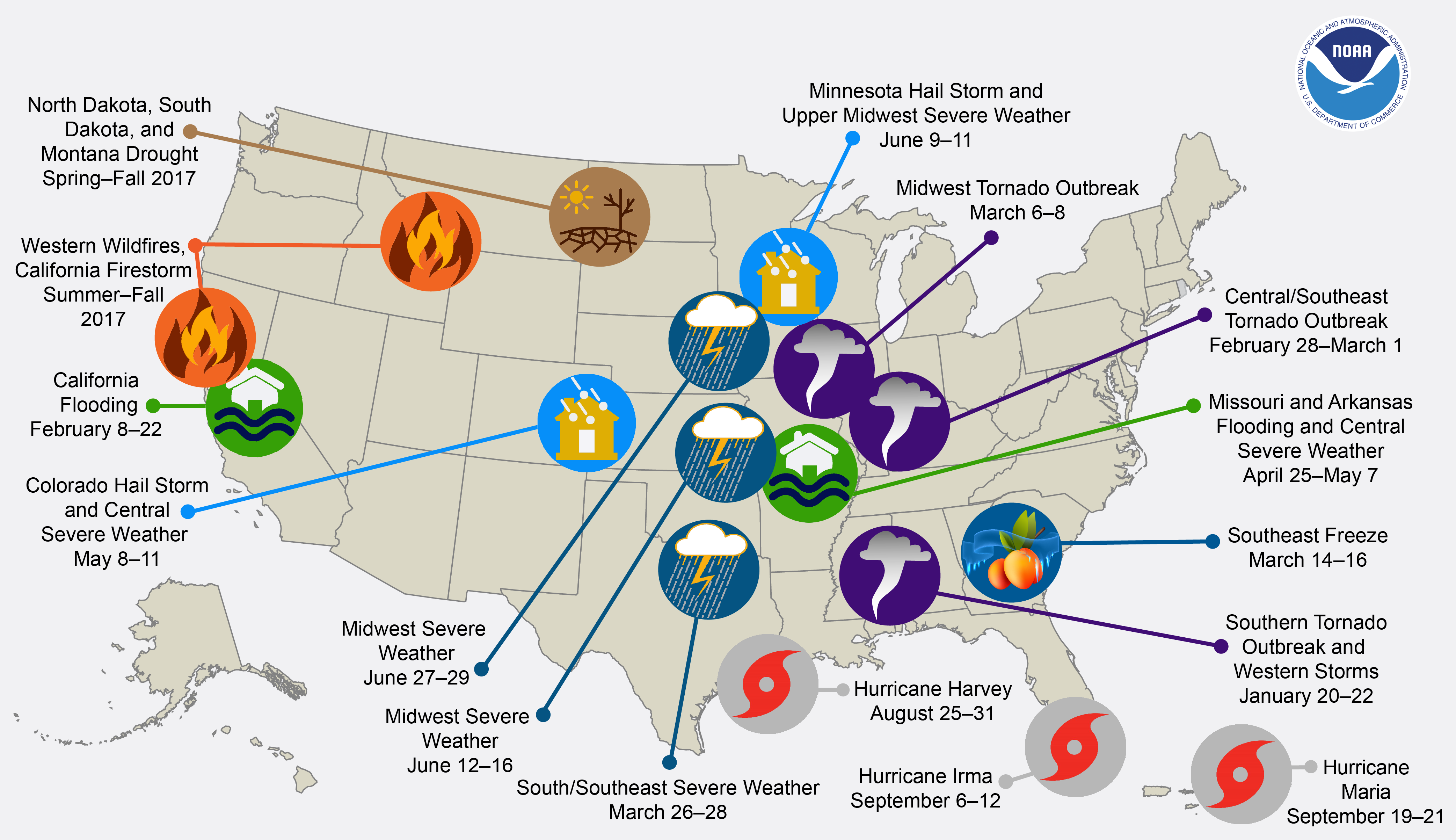}
\caption{U.S. weather-related disasters in 2017~\cite{NOAA-link}.}
\label{fig:disaster-map-2017}
\end{figure}

Natural disasters have always posed a critical threat to human beings, often being accompanied by major loss of life and property damage. In recent years, we have witnessed more frequent and intense natural disasters all over the world. In 2017 alone, there were multiple devastating natural disasters, each resulting in hundreds of deaths. Hurricanes, flooding, tornadoes, earthquakes and wildfires, were all active keywords in 2017. 
%Hurricanes Harvey, Irma, and Maria swept through the Atlantic in August and September. 
% A massive magnitude 7.1 earthquake shook Mexico City and killed 369 people. In an earthquake on the Iraq-Iran border in November, the number of death reached 620 and more than 8,000 were injured. The worst floods in a decade poured over South Asia: 1,400 people were killed in monsoon rains that hit India, Nepal, and Bangladesh~\footnote{http://www.businessinsider.com/worst-natural-disasters-hurricane-flood-wildfire-2017-12}. The wildfires that raced across California in 2017 caused unprecedented levels of death and destruction. 
An illustration of the distribution of weather-related disasters in a single year in the U.S. is presented in Figure~\ref{fig:disaster-map-2017}. To mitigate the impacts of disasters, it is important to rapidly match the available rescue resources with disaster victims who need help in the most efficient way, in order to maximize the impact of the rescue effort with limited resources. A key challenge in disaster rescues is to balance the requests for help with the volunteers available to meet that demand.

\begin{figure}[t]
 \centering
 \includegraphics[width=0.73\linewidth]{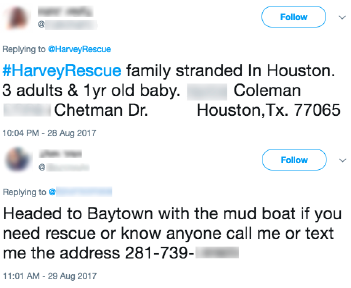}
 \caption{\label{fig:needs_example} Sample tweets requesting rescue and offering help.}
\end{figure}

\begin{figure}
    \centering
        \includegraphics[width=0.6\linewidth]{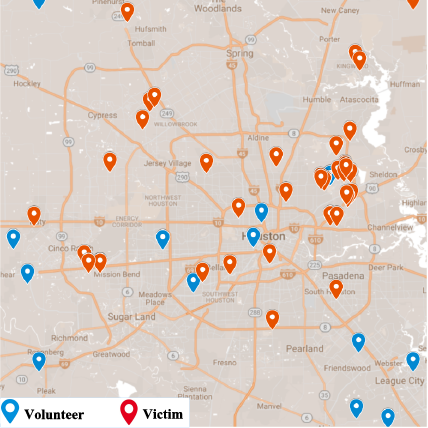}
    \caption{The distribution of volunteers and victims in the Houston area on August 28, 2017.}
    \label{fig:vv_distribution}
\end{figure}

The adverse impacts of a disaster can be substantially mitigated if during the disaster accurate information regarding the available volunteers can be gathered and victims' locations can be determined in a timely manner, enabling a well-coordinated and efficient response. 
This is particularly apparent whenever there is a huge burst of requests for limited public resources. For example, when Hurricane Harvey made landfall on August 25, 2017, flooding parts of Houston, the 911 service was overwhelmed by thousands of calls from victims in a very short period. Since the phone line resource is limited, many phone calls did not get through and victims turned to social media to plead for help, posting requests with their addresses. At the same time, many willing volunteers seeking to offer help during the disaster were left idle as no one knew where they should be sent.
This case is illustrated in Figure~\ref{fig:needs_example}, along with a sample distribution of victims and volunteers in Figure~\ref{fig:vv_distribution}. In the case of a hurricane, a major challenge is that without coordination, multiple volunteers with boats may go to rescue the same victim while other victims have to wait for extended times to be rescued.
This mismatch between victims and volunteers represents an enormous waste of limited volunteer resources. It is therefore imperative to improve the emergency services' coordination to enable them to efficiently share information, coordinate rescue efforts and allocate resources more effectively, and offer guidance for optimal resource allocation.

The problem of resource coordination has drawn considerable attention in the computer science community, and several data mining frameworks have been developed to address this problem. 
%Previous researchers have tended to focus on three approaches: supervised learning, adaptive methods, and heuristics. \textcolor{red}{We are also heuristics!}\textcolor{orange}{I think the traditional heuristic is different, since they use only heuristic }
Previous researchers have primarily focused on three approaches: supervised learning, adaptive methods, and optimization-based method.
Traditional supervised learning models demand a dataset that is statistically large in order to train a reliable model~\cite{Zhu06semi-supervisedlearning}, for example by building regression models to predict needs and schedule resources accordingly~\cite{5691343,ISLAM2012155}. %However, regression model heavily relies on the consistency of data and the timeliness of prediction. %Thus the performance is unstable.
Unfortunately, due to the unique nature of resource management for disaster relief, it is generally impractical to model this using traditional supervised learning models. Every disaster is unique and hence it makes no sense to model one disaster relief problem by using the dataset collected from other disasters; a realistic dataset for that disaster can only be obtained when it occurs. This means that traditional supervised learning is unable to solve the highly individual resource management problems associated with disaster relief efforts. 

%Some works focus on resource provisioning with budget constraints~\cite{Zhu:2010:RPB:1851476.1851516}. 
Other researchers have developed adaptive methods~\cite{6565979,IQBAL2011871} and proposed adaptive systems~\cite{6137322} for resource allocation. However, a common limitation of the adaptive approach is that the parameters in adaptive models change slowly and hence converge slowly. %\textcolor{red}{How about our method? converge slowly?}
An alternative is to model resource coordination problems as simulation problems or optimization problems which requires the process of modelling and tuning repeatedly if any of the external environmental parameters change. 

Real world resource coordinating problems are very challenging for a number of reasons:
\begin{enumerate}
      \item The sample size is small, especially in the early stages of the disaster, when there is almost no available data. Any decision-support system needs to move fast and make decisions swiftly. 
      \item The real-world environment where the resource coordination actually happens is a highly complex system with multiple uncertainties. For instance, the locations of volunteers and victims are dynamically changing, and the rescue time for an arbitrary victim varies depending on factors such as traffic, road closures, and emergency medical care, many of which are also changing dynamically. 
      \item There is no well-defined objective function to model the scheduling problem for disasters, especially when victims need emergency care or collaborative rescue efforts. 
\end{enumerate}

      %~\cite{Zhang:2017:TOD:3097983.3098138}.
      %\item The underlying system may change, and hence the inputs might vary greatly.
      
      %Since we can only estimate some parameters to approximate the real-world environment, it is difficult to depict the underlying system accurately with a mathematical model. 
      
    %to find the appropriate performance metrics and object function in some cases~\cite{Zhang:2017:TOD:3097983.3098138}.

%Because of the difficulties mentioned above, we seek alternative approaches to solve the resource management problem. 
The recent success achieved in applying machine learning to challenging decision-making domains~\cite{DBLP:journals/corr/MnihKSGAWR13,44806,DBLP:journals/corr/abs-1708-05866} suggests that Reinforcement Learning (RL) is a promising method with considerable potential. So far, reinforcement learning has been successfully applied to solve problems such as optimizing deep neural networks with asynchronous gradient descents for the  controllers~\cite{DBLP:journals/corr/MnihBMGLHSK16}, playing Atari with reinforcement learning~\cite{DBLP:journals/corr/MnihKSGAWR13}, learning control policies in a range of different environments with only very minimal prior knowledge~\cite{VolodymyrMnih2015Hctd,44806}, among others.
One appealing feature of the reinforcement learning method is that it can overcome many of the difficulties involved in building accurate models, which is usually formidable given the scale and complexity of real-world problems. Moreover, reinforcement learning does not require any prior knowledge of system behavior to learn optimal strategies. This means that reinforcement learning can be used to model systems that include changes and/or uncertainties. Finally, reinforcement learning can be trained for objectives that are hard to optimize directly because of the lack of precise models. When reward signals that are correlated with the objective are involved, this can be modelled by reinforcement learning as it is possible to incorporate a variety of goals by adopting different reinforcement rewards. 

In this paper, we aim to find an effective way to coordinate the efforts of volunteers and enable them to reach disaster victims as soon as possible.  
We have developed a novel heuristic multi-agent reinforcement learning-based framework to analyze the tweets and identify volunteers and victims, along with their locations.
Based on the information collected, a resource coordination system can then allocate the volunteer resources more efficiently. The resource coordination system is implemented using heuristic multi-agent reinforcement learning since this approach offers a good way to address the above dilemmas because of its unique characteristics. More specifically:
\begin{itemize}
    \item We build an efficient heuristic multi-agent reinforcement learning framework for large-scale disaster rescue work based on information gathered by mining social media data. This study is one of the first that specifically focuses on coordinating volunteers in disaster relief using reinforcement learning.
    \item We propose a \theName{} algorithm,
    %based on heuristic multi-agent reinforcement learning,
    which is capable of adapting dynamically as information comes in about volunteers and victims' situations and makes recommendations to minimize the total distance travelled by all the volunteers to rescue the maximum possible number of victims.
    \item Our proposed new disaster relief framework bridges the gap when traditional emergency helplines such as 911 are overwhelmed, thus benefiting both the disaster victims and the non-Governmental organizations seeking to help them.
    \item Last but not least, our proposed \theName{} algorithm significantly outperforms existing state-of-the-art methods, reducing the computation times required considerably. The effectiveness of the proposed method is validated using a Hurricane Harvey related social media dataset collected in August 2017 for the Houston area, Texas.
\end{itemize}

\section{Related Work}
% We review four categories of related work as follows.
\paragraph{\textbf{Disaster Relief with Social Media.}}
The most recent survey~\cite{NazerTahora2017IDRv} pointed out that, the success of a disaster relief and response process relies on timely and accurate information regarding the status of the disaster, the surrounding environment, and the affected people. There are a large number of studies using social media data for disaster relief. Gao et al. ~\cite{gao2011harnessing} built a crowdsourcing platform to provide emergency services during the 2010 Haiti earthquake, such as handling food requests. They integrated the system with crisis maps to help organizations to identify the location where supplies are most needed. Zook et al. ~\cite{WMH3:WMH344} demonstrated that information technologies were the key means through which individuals could contribute to relief efforts without being physically present in Haiti. This example proved how to make full use of volunteer resources by outsourcing tasks to remote volunteers. Ashktorab et al.~\cite{DBLP:conf/iscram/AshktorabBNC14} introduced a Twitter-mining tool to extracts practical information for disaster relief workers during natural disasters. Their approach was validated with tweets collected from 12 different crises in the United States since 2016. In the work of~\cite{houston2015social}, they identified fifteen distinct disaster social media uses, ranging from preparing and receiving disaster warnings, detecting disasters before an event to (re)connecting community members following a disaster. Gao et al.~\cite{Gao:2013:ETE:2507157.2507182} proposed a model to explore users' check-in behaviors via Location-based social networks and they integrated users' check-in history information to help build the connections between historical records and predicted locations. Lu et al. ~\cite{lu2015visualizing} explored the underlying trends in positive and negative sentiment concerning disasters and geographically related sentiment using Twitter data.

\paragraph{\textbf{Multi-agent Reinforcement Learning}}
The research on Multi-agent Reinforcement Learning (MARL) has proved to be very challenging. 
% Numerous researchers have been working in this field, and some of the contributions and achievements help remove barriers to the successful implementation and application MARL. 
The exponential growth of the discrete state-action space gives rise to a challenge for iterating over the state-action space.
The correlated returns of multiple agents make it difficult to maximize the returns independently. Several MARL goals have been proposed to circumvent this problem. Hu and Wellman proposed a framework where agents maintain Q-functions over joint actions and perform updates based on agents' learning dynamics~\cite{Hu03nashq-learning}. Powers and Shoham proposed to consider the adaption to the changing behaviors of the other agents~\cite{Powers05newcriteria}. Other researchers also proposed to consider both stability and adaption at the same time~\cite{NIPS2004_2673,Bowling:2001:RCL:1642194.1642231,LITTMAN200155}.
Nonstationarity arises in MARL since all the agents in the system are learning simultaneously. Stability essentially means the convergence to a stationary policy, whereas adaptation ensures that performance is maintained or improved when the other agents are changing their policies~\cite{Buşoniu2010}. It has been discovered that convergence is required for stability and rationality is the criterion for adaption~\cite{Bowling:2001:RCL:1642194.1642231}. An alternative to rationality is the concept of no-regret, which prevents the learner from being `exploited' by other agents~\cite{NIPS2004_2673}. Convergence to equilibria is a basic stability requirement~\cite{Hu03nashq-learning}, and it means that agents' strategies should eventually converge to a coordinated equilibrium. Nash equilibrium is commonly used in some scenarios, but concerns have been raised regarding its usefulness\cite{SHOHAM2007365}.

\paragraph{\textbf{Resource Management}}
% The resource management problem is of great importance not only in computer systems and networks but also in other areas, such as rescue systems. 
Studies of resource management problems appear in different fields, including real-time scheduling in CPU~\cite{Liu:1973:SAM:321738.321743,Audsley91real-timescheduling:}, energy resource efficiency management of data center for cloud computing~\cite{Beloglazov:2012:ERA:2148243.2148369,Beloglazov:2010:EER:1844765.1845139}, bitrate adaptation in video streaming~\cite{Jiang:2012:IFE:2413176.2413189,Huang:2014:BAR:2619239.2626296}, network congestion controls~\cite{CHIU19891}, and so on. Tesauro et al. combined reinforcement learning and queuing models to learn resource valuation estimates, and results showed their model achieved significant performance improvements over a variety of initial model-based policies~\cite{Tesauro06ahybrid}. Xu et al. presented a unified reinforcement learning approach to automate the configuration of virtualized machines and appliances running in the virtual machines. Their model achieved an optimal or near-optimal configuration setting in a few trial-and-error iterations~\cite{XU201295}. Mao et al. built a resource management system with reinforcement learning. Their initial results show that model based on deep learning outperforms the heuristic model, and their model can adapt to different conditions, converges quickly, and learns strategies that are sensible in hindsight~\cite{Mao:2016:RMD:3005745.3005750}.

% \section{Volunteer Resource Routing with Reinforcement Learning} 
\section{Problem Formulation}
We first introduce a formal definition of the rescue problem, then present the concept of multi-agent reinforcement learning and discuss its applications to disaster relief, concluding by presenting our new methodology. 

%\subsection{Optimization-based Problem Formulation}
%\vspace{-4mm}
\subsection{Problem Formulation}
\begin{definition}[Volunteering rescuer]
A volunteer rescuer (or volunteer) is a person who has access to potential rescue facilities (e.g. a boat) and is willing to help. We define all the volunteers at time $t$ as a set $U_t=\{u_1,u_2,...u_{M_t}\}$.
\end{definition}

\begin{definition}[Victim]
A victim is an individual who is trapped or in trouble and needs to be rescued. We denote all the victims at time $t$ as a set $V_t=\{v_1,v_2,...v_{N_t}\}$.
\end{definition}

% \textcolor{red}{
%     \begin{definition}[Obstacle]
%     An obstacle is a thing that blocks the way that volunteer goes to rescue a victim. It could be a big hole, fallen tree across the street, etc.. We denote this as a set $O=\{o_1,o_2,...o_k\}$.
%     \end{definition}
% }

\begin{definition}[Rescue Task]
Let $U_t=\{u_1,u_2,...u_{M_t}\}$, and $V_t=\{v_1,v_2,...v_{N_t}\}$ denote the volunteers and victims, respectively. These volunteers and victims are scattered across certain areas affected by the disaster. A rescue task is to find a volunteer $u_i$ in the volunteer set to rescue a victim $v_j$ that is trapped at another location. The `cost' of such a rescue task is the total time that it takes for volunteer $u_i$ to reach victim $v_j$ and convey them to a safe place. 
\end{definition}

For the purpose of this study, we assume that victims are taken to the nearest shelter after they have been rescued. We calculate the total time for a rescue task as $T=T(D)_{travel}+T_{load}+T_{shelter}$, where $D$ is the distance between the volunteer and the victim, $T(D)_{travel}$ is the travel time that it takes for the volunteer to reach the victim, $T_{load}$ is the time to load the victim(s) to the boat, and $T_{shelter}$ is the time needed to carry them to the nearest shelter. Since the loading time $T_{load}$ and the time to shelter $T_{shelter}$ are constants in every scheduling policy, we 
will not take the loading time $T_{load}$ and the time to shelter $T_{shelter}$ into consideration.
%only consider and compare the travel time $T(D)_{travel}$.

% \textcolor{blue}{In this paper, we assume that a victim will be sent to the nearest shelter if there is available capacity, and we also assume the total capacity of all the shelters are larger than or equal to the number of victims. Also, for simplicity we assume that a victim can rescue all the victims mentioned in a rescue request. These assumptions are valid and reasonable in practical, and we treat them as the default settings and won't mention it in the experiment part.}

\begin{definition}[Rescue Scheduling Task]
Let $A_t$ denote the set of assignments of victims to be rescued by volunteers at time $t$. Given a set of volunteers $U_t=\{u_1,u_2,...u_{M_t}\}$, and a set of victims $V_t=\{v_1,v_2,...v_{N_t}\}$, a rescue scheduling task is to find a set of sequential assignments of volunteers to rescue victims, such that all the victims are rescued with minimal total cost. The total cost for such scheduling is the total time spent on rescuing all the victims.
\end{definition}

Suppose that the cost function $C:X_t \to [0, \infty]$, where $X_t \in A_t$ is an assignment of volunteers for rescuing. The cost function is interpreted as a ``total rescuing time", and can be expressed in terms of times ${C_t}:U_t \times V_t \to  [0, + \infty]$, the cost/time for volunteers $U_{t}$ to rescue victims $V_{t}$. The rescue scheduling problem is to find the optimum assignment $X_t \in A_t$ such that $C_t(X_t)$ is a minimum, that is, there is no assignment $Y_t \in A_t$ such that $C(X_t) > C(Y_t)$.

Assignment $X_t \in A_t$ may be written as an $N_t \times M_t$ matrix, in which column $i$ lists the victims that volunteer $U_i$ will rescue at time $t$, in order. Suppose there are $N_t$ victims to be rescued by $M_t$ volunteers. We can now represent the rescue scheduling result as a matrix $X_t = (x_{ij})_{{N_t}{M_t}}$
% \begin{equation}
%     \nonumber
%     X=\left( {
%         \begin{array}{*{20}{c}}
%         {{x_{11}}}& \cdots &{{x_{1M}}}\\
%          \vdots &{{x_{ij}}}& \vdots \\
%         {{x_{N1}}}& \cdots &{{x_{NM}}}\\
%         \end{array}} 
%     \right)
% \end{equation}

\begin{equation}
    \nonumber
  x_{ij}=
  \begin{cases}
    1 & \text{volunteer i is dispatched to rescue victim j,} \\
    0 & \text{volunteer i is not dispatched to rescue victim j.}
  \end{cases}
\end{equation}
where $1 \leq i  \leq N$, $1 \leq j  \leq M$.

In this case, a volunteer rescues one victim at a time, while a victim can only be rescued by at least one volunteer. The mathematical model for the volunteer-victim problem is defined as follows:

\begin{equation*}
\begin{aligned}
& \underset{x}{\text{minimize}}
& & 
C = \sum\limits_{t = 1}^T {\sum\limits_{i = 1}^{{N_t}} {\sum\limits_{j = 1}^{{M_t}} {{d_{ij}}{x_{ij}}} } } \\
& \text{subject to}
& & \sum\limits_{i=1}^{N_t}{x_{ij} }\leq 1, \; j = 1, \ldots, M_t;\\
&&&\sum\limits_{j=1}^{M_t}{x_{ij} }\geq 1, \; i = 1, \ldots, N_t;\\
&&&{x_{ij}} \in \{ 0,1\}.
\end{aligned}
\end{equation*}
where ${d_{ij}}$ is the distance from volunteer i to victim j. 
% \begin{theorem}[NP-Hard]
% The optimization problem to find the optimal total time of scheduling $M$ volunteers (denoted as $U=\{u_1,u_2,...u_M\}$) to rescue $N$ victims (denoted as $V=\{v_1,v_2,...v_N\}$) is NP-hard.
% \end{theorem}

% It has been proved that the scheduling problem is NP-hard~\cite{lenstra1981complexity}. \textcolor{red}{
% Traditionally, these problems can be solved approximately by well-designed heuristic methods.}
% However, the process of modeling and seeking solutions for the model is inflexible, time-consuming and expensive. Moreover, the process of modeling and tuning up a heuristic method to achieve a decent performance usually has to be repeated if one or more environmental parameters change. This challenge is addressed by
% introducing multi-agent reinforcement learning (MARL).

\subsection{\theName{}: Heuristic Multi-agent Reinforcement Learning (MARL) in Rescue Scheduling}
\paragraph{\textbf{The setting of MARL}}
To tackle this rescue scheduling problem, we can formulate the problem using multi-agent reinforcement learning technique\cite{littman1994markov}. The agents are volunteers who are willing to rescue disaster victims. The victims represent the rewards and the environment is the place where the disaster happened. This environment is represented as a square-grid world, and the agents move within this grid world to rescue the victims. In other words, this is a Markov game G for N agents , which is denoted by a tuple $G=<\rm N,S, A, P, R, \gamma>$, where $N$, $S$, $A$, $P$, $R$, $\gamma$ are the number of agents, sets of states, joint action space, transition probability function, reward function and discount factor respectively. These are defined as follows:

\begin{itemize}
    \item \textbf{Agent}: We consider a volunteer with a boat to be an agent. Although the number of unique heterogeneous agents is always $N$ , the number of agents $N_t$ is changing over time t.
    \item \textbf{State $s_t \in S$}: A state $s_t^i$ of a volunteer $i$ at time $t$ in the rescue scheduling problem is defined as the possible grid location where he or she is located. We also maintain a global state $s_t$ at each time $t$, considering the spatial distributions of available volunteer and victims as a global state $s_t \in S$, and the states $S$ is a finite set.
    \item \textbf{Action $a_t \in A $= $A_1 \times \ldots \times A_{N_t}$}: a joint action $a_t=\{a_t^i\}_1^{N_t}$ denotes the allocation strategy of all available volunteers at time $t$, where $N_t$ is the number of available agent at time $t$. The action space $A_i$ of an agent $i$ specifies all the possible directions of motion for the next iteration, which gives a set of four discrete actions denoted by ${k}_{k=1}^4$ represented for $\{up, down, right, left\}$ transition; allocating the agent to one of the four neighboring grids. At time $t$, if the state $s_t^i$ and the action $a_t^i$ of an agent is given, then we conclude its state $s_{t+1}^i$ at time $t+1$. Furthermore, the action space of agents depends on their locations. The agents located at corner grids have a smaller action space. 
    %We also assume that the action is deterministic: if $a_i^t = [g_0, g_1]$, then agent $i$ will arrive at the grid $g_1$at time $t + 1$. 
    A policy $\pi$ is defined as a sequence of actions and ${\pi}^*$ is the optimal policy.
    
    \item \textbf{Discount Factor $\gamma$}: A discount factor $γ$ is between 0 and 1, and it is used to quantify the difference in importance between immediate rewards and future rewards.
    
    \item \textbf{Transition function $P: S \times A \to [0,1]$}: Transition function gives the probability that describes the probabilities of moving between states. The state transition probability $p(s_{t+1}|s_t, a_t)$ gives the probability of transiting to $s_{t +1} \in S$ given a joint action $a_t \in A_i$ is taken in the current state $s_t \in S$. 
    
    \item \textbf{Reward function $R_i \in R = S \times A \to (-\infty, +\infty)$}: A reward in the rescue scheduling problem is defined as the feedback from the environment when a volunteer takes actions at a state. Each agent is associated with a reward function $R_i$ and all agents in the same location have the same reward function. The $i-th$ agent attempts to maximize its own expected discounted reward:
    ${R_t}=E({r_t^i}+{\gamma}{r_{t+1}^i}+...)=E({\sum\limits_{k=0}^{\infty}{\gamma}^k}{r_{t+k}^i})=E({r_t^i}+{\gamma}{R_{t+1}})$. 
    % The individual reward $r_t^i$ for the i-th agent is associated with the action $a_t^i$ is defined as the averaged revenue for all agents arriving at the same grid as the i-th agent at time $t+1$.
\end{itemize}

The goal of our disaster rescuing problem is to find the optimal policy $\pi^*$ (a sequence of actions for agents) that maximizes the total reward. The state value function ${V^\pi}(s)$ is introduced to evaluate the performance of different policies. ${V^\pi}(s)$ stands for the expected total reward with discount from current state $s$ on-wards with the policy $\pi$, which is equal to:

\begin{equation}
\begin{split}
{V^{\pi}}(s) & = {E_{\pi}}({R_t}|S={s_t}) = {E_{\pi}}({r_t}+{\gamma}{V^{\pi}}({s'}))\\
 & = {r_t}+{{\sum\limits_{{s'} \in S}}{P^{\pi}}({s'}|s){V^{\pi}}({s'})}.
\end{split}
\end{equation}

According to Bellman optimality equation~\cite{Bellman:1957}, we have \begin{align}{V^\pi}(s)={\max_{a \in A}}{r_t}(s,a)+{\sum\limits_{{s'} \in S}}{\gamma}{P^{\pi}}({s'}|s,a){V^\pi}({s'}).\end{align} %{Similarly, $Q$-function, the expected reward given the agent taking action $a$ in state $s$ and following policy $\pi$, is defined as \begin{align}{Q^{\pi}}(s,a)=E\{{\sum\limits_{k=0}^\infty}{\gamma^k}{r_{t+k+1}}|{s_t}=s, {a_t}=a, \pi\}.\end{align}}

Since the volunteers have to explore the environment in order to find victims, they cannot observe the underlying state of the environment. We treat this as a Partially Observable Markov Decision Process (POMDP)\cite{Puterman:1994:MDP:528623}. A POMDP extends the definition of Markov Decision Process (MDP). It is defined by a set of states $S$ denoting the environment setting for all agents, a set of actions $A_1...A_N$ and a set of observations $O_1...O_N$ for each agent. The state transition function $P:S \times {\rm A_1} \times...\times {\rm A_N} \to {\rm S}$ produces the next state with agents taking the action following the policies ${\pi_{\theta_i}}: {\rm O} \times {A_i} \to [0,1]$. Each agent $i$ receives an observation correlated to the state ${o_i}: {\rm S} \to {\rm O_i}$, and obtains a reward ${r_i}: {\rm S} \times {\rm A_i} \to {\rm R}$. Each agent $i$ aims to maximize the shared total expected return ${R_i}=\sum\limits_{i=1}^{N}{\sum\limits_{t=0}^T}\gamma^{t}{r_i^t}$ where ${\gamma}$ is the discount factor and $T$ is the horizon.

% \begin{algorithm}[t]
%     Initialize the Q-value for volunteers to arbitrary $Q(s,a)$\;
%     episode = 0\;
%     \Repeat{episode > TOTAL\_EPISODES}{
%         episode =  episode + 1 \;
%         \Repeat{rescue complete}{
        
%             \For{volunteer\_i in Volunteers}{
%                 Observe current state $s$\;
%                 Select an action $a$ according to policy $\pi(s,a)$\;
%                 Execute action $a$\;
%                 Record reward ${r_t}$ and next state $s'$\;
%             }
            
%             ${Q({s_t},{a_t})}\leftarrow (1-\alpha){\cdot}Q({s_t},{a_t})+{\alpha}{\cdot}({r_t}+{\gamma}{\cdot}{max_{a}}Q({s_t+1},{a_t})$\;
        
%         }
%     }    
% \caption{Q-learning in rescue scheduling}
% \label{algorithm:Q-learning}
% \end{algorithm}

%**********new algorithm***********
\begin{algorithm}[t]
    % Initialize:\;
        let t=0, ${Q_t^i}$=1\;
        initialize $s_0$\;

    \Repeat{rescue complete}{
        % Observe current state $s_n$\;
        % Choose best actions $a_t^i$ based on policy ${\pi}^{i}(s_t)$\;
        % Every volunteer performs its action $a_t^i$\;
        % Observe reward $R_t^i$...$R_t$ and next state $a_t^i$...$_t$ \\
        % update policy ${\pi}^{i}(s_t)$\;
             
    % \For{volunteer $a_i$ in $A_i$}{
        Observe current state $S_t$\;
        $A_t$ = HeuristicActionSelection($S_t$) \\
        Every volunteer execute its action $a_t^i$ in $A_t$\;
        Observe $R_t^i$...$R_t$ and $a_t^i$...$_t$ \\
        % // update q table\\
        \[Q_{t + 1}^i(s,{a^1}...{a^N}) = (1 - {a_t}){Q^i}(s,{a^1}...{a^N}) + {a_t}\{ r_t^i + \]\[\gamma {\pi ^i}({s_{t + 1}})\sum\limits_{j = 1}^N {Q_t^j({s_{_{t + 1}}}){\pi ^j}({s_{t + 1}})} \} \]
         where $({\pi}^{i}(s_{t+1}), {\pi}^{j}(s_{t+1}))$ are cooperative strategies\;
        Let t=t+1\;
    % }
    
    }
\caption{\theName{} in Rescue Scheduling}
\label{algorithm:Q-learning}
\end{algorithm}

\begin{algorithm}[t]
    
    \SetKwInOut{Input}{Input}
    \SetKwInOut{Output}{Output}
    
    \underline{function HeuristicActionSelection} $(S_t)$\\
    \Input{State $S_t$}
    \Output{best $found\_action$}
    
    Choose best actions $A$ based on policy ${\pi}(S_t)$ and $Q$ \\
    ${min\_distance}$ = $\infty$ \\   
    
    \For{$action_n \in A$}{
        $next\_state_n$ = $perform\_actions(action\_n)$ \\
        $distance$ = HeuristicDistance($next\_state_n$) \\
        \If{$distance \leq min\_distance$}{
            $min\_distance$ = $distance$ \\
            $found\_action$ = $action_n$ \\
        }
    }
    \Return $found\_action$
\caption{Heuristic action selection}
\label{algorithm:heuristic-selection}
\end{algorithm}

\begin{algorithm}[t]

    \SetKwInOut{Input}{Input}
    \SetKwInOut{Output}{Output}

    \underline{function HeuristicDistance} $(S)$\\
    \Input{Current state $S$}
    \Output{heuristic distance at state $S$}

     - compute distances from agents to victims\\
     - sort distances in ascending \\
     - pick pair matching agent to the shortest victim\\
     - total\_distance = sum distance from agents to selected victims \\
     
     \Return total\_distance
    
\caption{Heuristic distance calculation}
\label{algorithm:heuristic-distance}
\end{algorithm}
%**********new algorithm***********

Several reinforcement learning algorithms have been proposed to estimate the value of an action in various contexts. These include the Q-learning, SARSA, and policy gradient algorithm. Among them, the model-free Q-learning algorithm stands out for its simplicity. In Q-learning, the algorithm uses a Q-function to calculate the total reward, defined as $Q: S\times A \to R$. Q-learning iteratively evaluates the optimal Q-value function using backups:
\begin{equation}
Q(s,a)=Q(s,a)+\alpha[r+\gamma{max_{a'}}Q(s',a')-Q(s,a)]
\end{equation}
where $\alpha \in [0,1)$ is the learning rate and the term in the brackets is the temporal-difference (TD) error. Convergence to $Q^{\pi^*}$ is guaranteed in the tabular case provided there is sufficient state/action space exploration.

\paragraph{\textbf{The heuristic function}}
The Q-learning requires a number of trials in order to learn and perform consistently, which will increase the total time to generate a rescue plan. In order to address this problem, heuristic-based algorithms have been proposed, e.g. in Robotic Soccer game \cite{bianchi2007heuristic}. For the current problem, we propose a heuristic based Q-learning: \theName{}. In our problem, the locations of volunteers and victims will be estimated via tweets' geolocation as described in Section \ref{geocoding-session}. We will then incorporate this information as a heuristics function in the learning process. When determining actions for volunteers, besides choosing the optimal Q-value as mentioned earlier, we also prioritize the actions that result in the shortest distance to the victims. The heuristics function is a mapping $H : S \times A \to R$ where $S$ is the current state, $A$ is the action to be performed, and $R$ is a real number representing the distance of volunteers to the victims. If after performing an action $a$ in $A$, the agent is at row $r_a$ and column $c_a$ of the grid, and its goal is the victim positioned at row $r_v$ and column $c_v$, then the heuristic distance $h$ is calculated as:

\begin{equation}
    h = |r_a - r_v| + |c_a-c_v|
\end{equation}

Our proposed \theName{} algorithm for rescue activity is illustrated in Algorithm~\ref{algorithm:Q-learning}, \ref{algorithm:heuristic-selection} and \ref{algorithm:heuristic-distance}.

\section{Experiments}
In this section, we show the experimental study using real data collected from Twitter to fully evaluate the performance of our proposed algorithm. We first introduce the dataset and data processing, then show how the identified volunteers, victims and locations are mapped into the reinforcement learning environment. Finally we evaluate the performance of the proposed \theName{} algorithm, and compare the performances achieved using traditional search-based methods.

%Since the experiment involves in complex intermediate experiments, such as natural language processing and short text classification, we also detail the experiment result and process of intermediate experiment. 
\subsection{Datasets}
\begin{figure}[t]
     \centering
     \includegraphics[width=\linewidth]{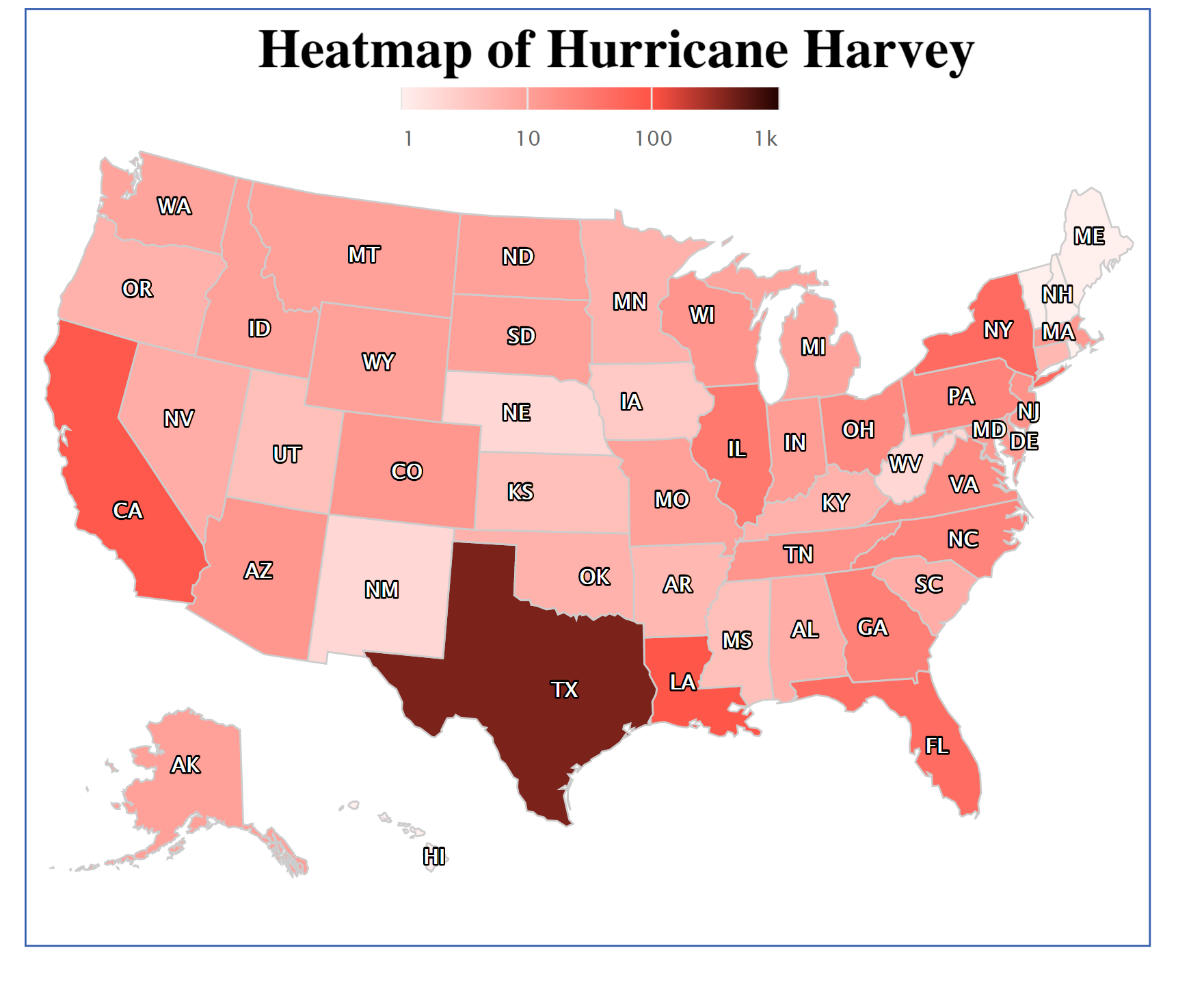}
     \caption{The geographical distribution of Hurricane Harvey tweets. This figure incorporates all the Hurricane Harvey-related tweets from Aug 17 to Sept 5, 2017. The total number of tweets are scaled to the range of 1 to 1000.}
     \label{fig:heatmap}
\end{figure}

%This paper presents a novel framework to identify victim's location and respond to victims' request in disasters, along with a case study using Twitter data in the recent Hurricane Harvey. This is one of the first attempts towards scheduling victims' request in real-time using massive social network data and reinforcement learning. A set of classifiers were trained to extract information about victims, volunteers and victims' needs in a disaster. Factors such as rescue plan, victim's needs, and emergency are also captured and fed into our scheduling engine. Our key contribution is designing a multi-agent reinforcement learning scheduling policy to simultaneously schedule multiple volunteers to rescue a group of victims effectively. The Multi-agent Reinforcement Learning can respond to dynamic requests and services and achieve good global performance. We test the system's performance with Harvey related tweets and prove our proposed approach outperform random walk and greedy algorithms.%, which are the two common scheduling methods in real disasters. %This system can be extended to other disaster relief work easily.
%We develop a tool to collect data from Twitter using Twitter API. 
Tweets were collected from Aug 23, 2017, to Sept 5, 2017 using Twitter API, covering the whole course of Hurricane Harvey and its immediate aftermath. The raw data for each day includes about two million tweets, each of which has 36 attributes including, among other information, the location of the tweet originated from, its geographic coordinates, and the user profile location. The raw data was cleaned by removing tweets that did not originate from the United States. Figure~\ref{fig:heatmap} shows a heat map of the Hurricane Harvey related tweets from Aug 23 to Sept 5, 2017. Not surprisingly, the state of Texas has the largest total number of tweets: 173,315.

% Table generated by Excel2LaTeX from sheet 'Sheet2'
\begin{table}[htbp]
  \centering
  \caption{Tweet distribution from Aug 23, 2017, to Sept 5, 2017.}
    \begin{tabular}{l|r|l|r}
    \toprule
    Total tweet & 25,945,502 & Volunteer tweet & 13,953 \\
    \midrule
    Harvey tweet & 173,315 & Victim tweet & 16,535 \\
    \bottomrule
    \end{tabular}%
  \label{tab:addlabel}%
\end{table}%

%This figure incorporates all the Hurricane Harvey-related tweets from Aug 17 to Sept 5, 2017.

% \begin{figure}[ht]
%     \centering
%     \includegraphics[width=\linewidth]{figures/TweetRate.png}
%     \caption{Tweeting rates relate to Hurricane Harvey and aid request. Tweeting rate is defined as the number of tweets posted in every second.}
%     \label{fig:statistics}
% \end{figure}

\begin{figure}[ht]
    \centering
    \includegraphics[width=\linewidth]{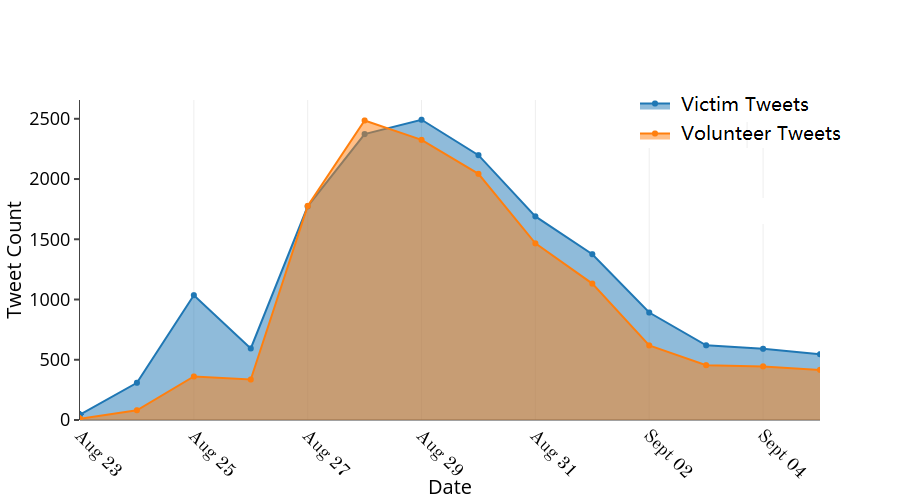}
    \caption{Time series of victim and volunteer tweet counts.}
    \label{fig:statistics}
\end{figure}

\setlength{\tabcolsep}{0.96pt}
\begin{table*}[t!]
    \caption{Tweets classification results. }
    \begin{center}
    \begin{tabular}{ccccccccccccc}
    \hline
        &\multicolumn{4}{c}{Harvey Classification} & \multicolumn{4}{c}{Victim Classification} & \multicolumn{4}{c}{Volunteer Classification}\\
        &\textbf{Precision} & \textbf{Recall} & \textbf{F\_ measure} & \textbf{Accuracy}  &\textbf{Precision} & \textbf{Recall} & \textbf{F\_ measure} & \textbf{Accuracy} &\textbf{Precision} & \textbf{Recall} & \textbf{F\_ measure} & \textbf{Accuracy} \\ [0.6ex]
    \hline
        Log. Regr. & 0.8 &  0.7273 &  0.7619 &  0.8646 & 0.8437 &  0.5510 &  0.6667 &  0.7127  & 0.9583 &  0.6216 &  0.7541 &  0.8170  \\

        KNN &  1.0  & 0.2105 &  0.3478 &  0.7580  & 1.0 & 0.8414 &  0.9139 & 0.9172 & 1.0  & 0.6129 &  0.76 &  0.8248\\

        CART &  1.0 &  0.6364 & 0.7778 &  0.8919 & 1.0 & 0.9795 & 0.9896 & 0.9893 & 1.0 &  0.7567 & 0.8645 &  0.8902\\

        SVM &  0.8947 &  0.9444 & 0.9189 &  0.9516 & 0.9146 & 0.9868 & 0.9493 &  0.9490 &  0.9146 &  0.9868 & 0.9493 &  0.9490 \\
    \hline
    \end{tabular}
    \label{table:classification}
    \end{center}
\end{table*}

%\subsection{Topic-Specific Tweet Classification}
\subsection{Identification of Victims and Volunteers}
\label{geocoding-session}
To identify victims of Hurricane Harvey and the volunteers wishing to help save them from social media, we fist designed a classifier to filter Harvey related tweets from all the collected tweets. In this context, a Harvey tweet refers to a post talking about Hurricane Harvey or related to Hurricane Harvey. Within the Harvey tweets, we further developed two classifiers to identify tweets from victims tweets and volunteers. Here, victim tweets are those from victims (or their friends) requesting help, including retweets. Volunteer tweets are those from volunteers who have boats and are willing to offer help. All three classifiers were implemented based on a Support Vector Machine (SVM). In every classifier, $2,000$ Harvey related tweets were manually labelled, with $80\%$ of the tweets being used for training, and the rest for testing. A five-fold cross-validation method was then applied to ensure the classification results were trustworthy. To obtain a reliable classification result, we compared Logistic Regression, K-Nearest Neighbor (KNN), CART, and SVM. 
Measurement criteria such as precision (positive predictive value), recall, F-measure, and accuracy were employed to measure the performance, as shown in Table~\ref{table:classification}.
% The performance is illustrated in Table~\ref{table:classification}.

\paragraph{Victim and volunteer time series}
% To monitor Hurricane Harvey rescue status, we tracked the Harvey-related tweet rate from Aug 17 to Sept 30, 2017, as shown in Figure~\ref{fig:statistics}. Here, the tweet rate means the number of tweets per second. Initially, when Hurricane Harvey began to form on Aug 17, not much attention can be extracted from Twitter in the US. As Hurricane Harvey approached Texas a week later, on Aug 24, there was a burst of Harvey related tweets. The tweet rate about Harvey peaked on August 25, 2017, while the tweet rate for requesting aid peaked on Aug 28, 2017. As the influence of Hurricane Harvey decreased, the tweet rates in all categories declined. 

To monitor the impact of Hurricane Harvey and rescue activities, we tracked the victims and volunteers tweets time series from Aug 23 to Sept 5, 2017, as shown in Figure~\ref{fig:statistics}. Initially, when Hurricane Harvey formed into a tropical depression on Aug 23, not much attention was observed from Twitter in the US. When Harvey made landfall near the Texas Gulf Coast on Aug 25, there was a burst of victims tweets. With the increasing of victims requesting help, number of volunteers also increased sharply and reached a climax on Aug 28. Meanwhile, victims tweets reached peak on Aug 29. With the leaving of Harvey and system-wide rescuing, both the victims tweets and volunteers tweets dropped gradually. Generally the number of volunteers tweets is always lower than the victims tweets. 

\paragraph{Geocoding}
To identify the victim and volunteer locations, we designed a simple tool based to extract the tweets' locations. 
For tweets from GPS-enabled devices that included geographic coordinates, or tweets giving specific addresses, we used the address directly to locate the victims or volunteers. 
Otherwise, we combined alternative sources of information to infer their location, such as the self-reported location string in the user's profile metadata, or by analyzing the tweet's content. With the help of the World Gazetteer (http://archive.is/srm8P) database, we were able to lookup location names and geographic coordinates.

% \begin{figure}[t]
%  \centering
%  \includegraphics[width=\linewidth]{figures/needs_flow.eps}
%  \caption{Needs flow. Different colours denote different needs, and the needs are presented on a daily basis according to the frequencies.}
%  \label{fig:needs}
% \end{figure}

% \setlength{\tabcolsep}{0.96pt}
% \begin{table*}[t!]
%     \caption{Tweets classification results. }
%     \begin{center}
%     \begin{tabular}{ccccccccccccc}
%     \hline
%         &\multicolumn{4}{c}{Harvey Classification} & \multicolumn{4}{c}{Victim Classification} & \multicolumn{4}{c}{Volunteer Classification}\\
%         &\textbf{Precision} & \textbf{Recall} & \textbf{F\_ measure} & \textbf{Accuracy}  &\textbf{Precision} & \textbf{Recall} & \textbf{F\_ measure} & \textbf{Accuracy} &\textbf{Precision} & \textbf{Recall} & \textbf{F\_ measure} & \textbf{Accuracy} \\ [0.6ex]
%     \hline
%         Log. Regr. & 0.8 &  0.7273 &  0.7619 &  0.8646 & 0.8437 &  0.5510 &  0.6667 &  0.7127  & 0.9583 &  0.6216 &  0.7541 &  0.8170  \\

%         KNN &  1.0  & 0.2105 &  0.3478 &  0.7580  & 1.0 & 0.8414 &  0.9139 & 0.9172 & 1.0  & 0.6129 &  0.76 &  0.8248\\

%         CART &  1.0 &  0.6364 & 0.7778 &  0.8919 & 1.0 & 0.9795 & 0.9896 & 0.9893 & 1.0 &  0.7567 & 0.8645 &  0.8902\\

%         SVM &  0.8947 &  0.9444 & 0.9189 &  0.9516 & 0.9146 & 0.9868 & 0.9493 &  0.9490 &  0.9146 &  0.9868 & 0.9493 &  0.9490 \\
%     \hline
%     \end{tabular}
%     \label{table:classification}
%     \end{center}
% \end{table*}

\begin{figure}[t]
     \centering
     \includegraphics[width=\linewidth]{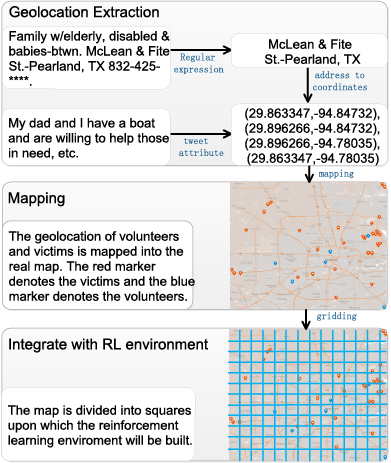}
     \caption{Reinforcement learning environment transformation.}
     \label{fig:expriment_geolocation}
\end{figure}

%\vspace{-2mm}
\subsection{Experiment Setting}
% \textcolor{red}{Need to specify how the heuristic works in the experiment setting, for example, how to define the heuristic function.}
% \textcolor{blue}{We mentioned it in the theory part.}
We can model the problem of rescue scheduling using a heuristic multi-agent fully cooperative reinforcement learning method. Multi-agent means that we use multiple agents to represent multiple volunteers. The number of agents depends on the number of volunteers identified in the volunteer tweet classification process for each day. Similarly, we assume that the victims are immobile learning targets due to the fact that victims are trapped. Since volunteers aim to rescue all the victims as soon possible, the goal of all agents is to reach all their targets with the lowest cost and maximize the total reward. %\textcolor{blue}{This is referred to as a fully cooperative agent interaction.}
% \textcolor{blue}{The interactions between agents might be either cooperative, competitive or mixed cooperative-competitive. In our case, we design that all the volunteers are fully cooperative to achieve the common goal, i.e., maximizing the shared total reward.}

\begin{figure*}[t]
	\centering
	\subfigure[Scatter Matrix Reward]{
	    \includegraphics[width= 7in,height=1.1in]{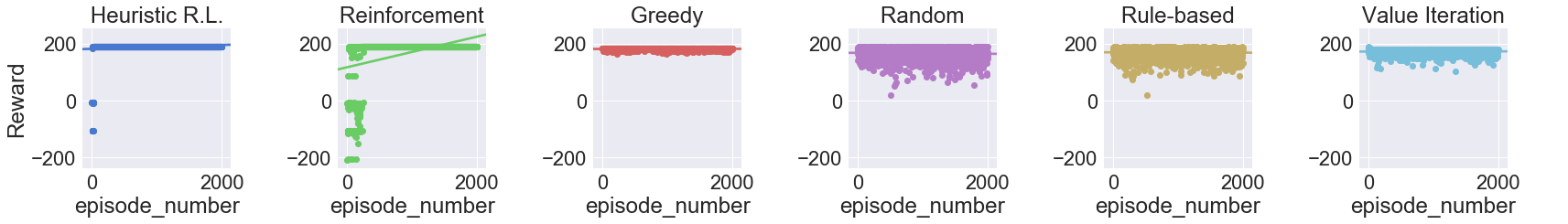}
	    \label{fig:expriment_score}
	}
    \subfigure[Scatter Matrix Time Step]{
		\includegraphics[width= 7in,height=1.1in]{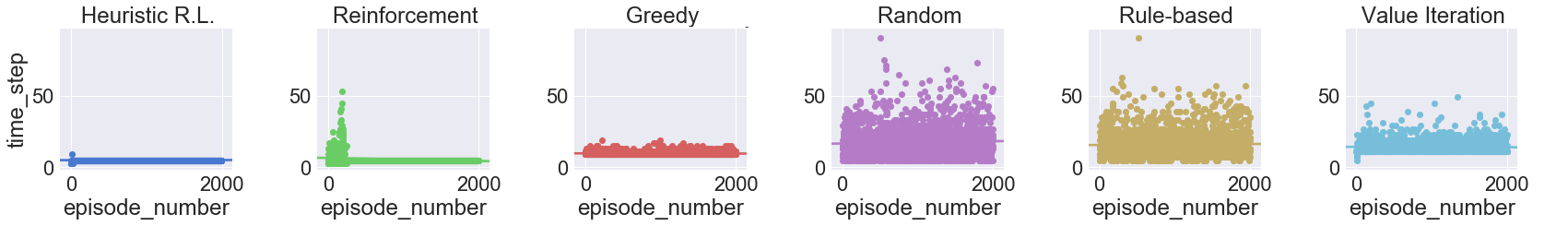}
		\label{fig:expriment_time}
    }
        \subfigure[Heatmap Reward]{
		\includegraphics[width= 3.3in,height=2in]{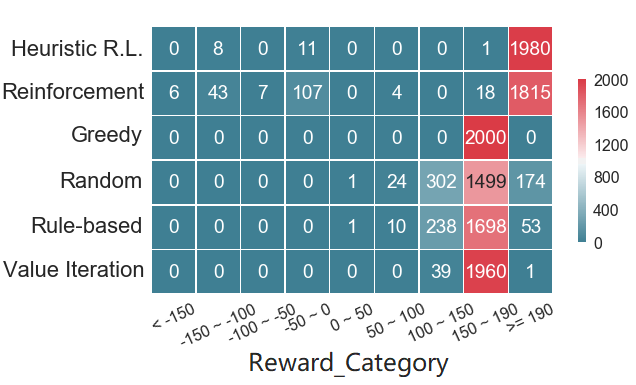}
		\label{fig:expriment_heatmap}
    }
        \subfigure[Box Plot Reward]{
		\includegraphics[width= 3.0in,height=2in]{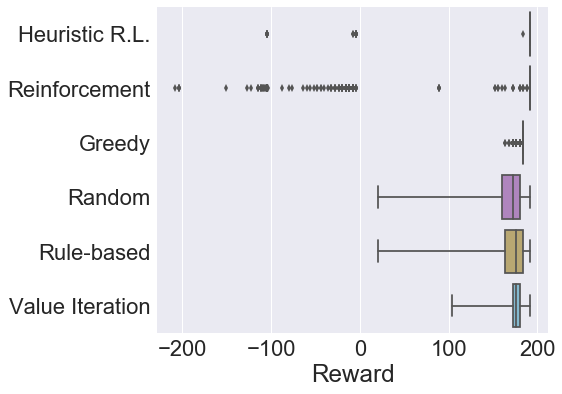}
		\label{fig:expriment_box_plot}
    }
\caption{Comparison of the performance of different algorithms.}
\label{fig:experiment}
\end{figure*}

In the following sections, we describe how the disaster grid environment is identified and what actions volunteers can perform in the course of their rescuing activities.

\subsubsection{The grid environment identification}
%Some transforming techniques are utilized to build the environment for the agents in our model. 
The process of environment building is illustrated in Figure~\ref{fig:expriment_geolocation}. In actual disaster relief operations, the whole city of Houston is the activity space for volunteers, and since a volunteer can go to any direction, the combination of space and direction will be infinite. According to our statistics, 95\% of the requests for help during the hurricane come from a fixed downtown area. For simplicity, our model is based on a quasi-square area defined by four position coordinates, which are (29.422486,-95.874178), (30.154665,-95.874178), (30.154665,-95.069705) and (29.422486,-95.069705), which are shown in Figure~\ref{fig:vv_distribution}. This square region has a width of 50 miles and for our purposes is mapped into a 25 by 25 grid, with each grid representing a 4-square-mile area in the real world. By applying this simple mapping to convert the actual map to a virtual grid, we can transform the real world continuous state space to a more manageable discrete state space, and hence significantly reduce the state space complexity.

The position coordinates of the victims and volunteers are extracted every hour following the processes described in the Figure~\ref{fig:expriment_geolocation}. This hourly updating strategy will keep our system updated with the number of available volunteers to be scheduled in order to go and rescue the remaining victims. From our observations, For victim tweets that contain the victim's address and phone number, such as \textit{McLean \& Fite St.-Pearland, TX 832-425-**** }, we can extract the address and converted their position coordinates. For volunteer tweets that do not include the volunteers' address, we can use the geocoding tool as described in section~\ref{geocoding-session} to extract geographical information from the raw tweet. This approximation is not precise but reasonable; as the volunteers will be moving around to rescue victims, their address is trivial. Similarly, every volunteer is mapped into the grid according to his/her position coordinates. 
%These coordinates are also the initial positions for volunteers and victims.

% \subsubsection{The action and state space}
% We defined four directions that volunteers can go in the grid environment. They are \textit{East, West, North} and \textit{South} directions. Moreover, it is possible that there are obstacles within the grid blocks the volunteers' direction.  

\subsection{Disaster Relief Coordination Performance}
\subsubsection{Baseline Models}
We used the following classical search methods to compare their performances with that of our proposed technique:

\paragraph{\textbf{Random walk}}
In this search policy, the agent will randomly walk surround the grid and search for any victim they come across along the way. The behavior is totally random without any other knowledge of the environment. 
\paragraph{\textbf{Greedy best first search}}
A greedy best first search offers volunteers a heuristic distance estimation to the victims. Volunteers begin by rescuing the closest victims first and then move on to the further ones sequentially. 
% \paragraph{\textbf{A* search}}
% An A* search also utilizes the provided heuristics related to distance estimation to the victims, as in the greedy approach. However, this method also takes into account of the total steps that the volunteer has already traveled.

% \begin{table*}[ht]
%     \setlength{\tabcolsep}{6pt}

%     \caption{Rescue performance comparison. Bold values represent best performance. \textcolor{red}{need to fill true result}}
%     \begin{center}
%     \begin{tabular}{| c | c | c | c | c | c | c | c | c |}
%     \hline
%         & \multicolumn{2}{| c |}{Episode Time} & \multicolumn{2}{| c |}{Episode Reward} & \multicolumn{2}{| c |}{Reward Rate}
%          & \multicolumn{2}{| c |}{Rescuing Cost} \\
%     \hline
%         & Average & Variance & Average & Variance & Average & Variance & Average & Variance \\ 
%     \hline
%     \hline

%         Random Walk &  17.4 &  0 &  167.2 &  0 & 9.6 & 0 & 0.104 &  0   \\

%         Greedy B.F.S. &  9.6   & 0 &  182.7 &  0  & 19.03 & 0 & 0.053 &  0 \\

%         A* Search &  9.0 &  0 & 184.0 &  0 & 29.44 & 0 & 0.034 &  0 \\

%         Rule-Based &  15.9 &  0 & 170.2 &  0 & 10.70 & 0 & 0.093 &  0 \\
%         Value Iteration &  14.1 &  0 & 173.7 &  0 & 12.32  & 0 & 0.081 &  0 \\

%     \hline
%     \hline
%         Proposed Method &  \textbf{5.0} &  0.0 & \textbf{192.0} &  0 & \textbf{38.4} & 0 & \textbf{0.026} &  0  \\
%     \hline
%     \end{tabular}
%     \label{table:performance-comparison}
%     \end{center}
% \end{table*}

\paragraph{\textbf{Rule-based search}}
A rule-based search computes action rules by utilizing the probability of taking an action in a grid cell. The action with highest probability are then selected for the next action. This probability is computed from the last average rewards gained in those cells during training episodes controlled by the random walk algorithm. In particular, if $V(t,j)$ is the averaged reward value at time $t$ of the grid cell $g_j$, and the volunteer takes action $a_t$ in order to move to grid cell $g_{j+1}$, the probability of taking action $a_t$ at the grid cell $g_j$ is:

\begin{equation}
    p(a_t=[g_j, g_{j+1}])= \frac{V(t+1, j)}{V(t+1,j) + V(t+1, j+1)}
\end{equation}

\paragraph{\textbf{Value iteration}}
This algorithm works by dynamically updating the value table based on a policy evaluation such as that described by \cite{sutton1998reinforcement}. The allocation policy is computed based on the new value table,

\paragraph{\textbf{Reinforcement Learning}}
This is traditional reinforcement learning technique where there is no heuristics consideration in action selection. The technique has the same settings such as action, state and reward space compared with our proposed heuristic reinforcement learning.

\subsubsection{Evaluation Metrics}
We define an episode as the set of attempts made by all volunteers to successfully rescue all the victims. Hence, our key metrics for measuring the performance of rescue activities are the \textit{average episode time}, \textit{average episode reward}, \textit{average reward rate} and \textit{average rescuing cost}. 

\paragraph{\textbf{Average episode time}} is the average total time steps required to rescue all the victims in all executed episodes. Each time step is equivalent to one step action (from one cell to an other near-by cell) taken by all the available volunteers.

\paragraph{\textbf{Average episode reward}} is the average cumulative reward that all volunteers earn in each episode of rescuing.

\paragraph{\textbf{Average reward rate}} is the ratio between the average episode reward and average episode time. This represents the average reward that all volunteers earn in one time step. If there are $N$ episodes, and $reward_i$ and $time_i$ represents the reward and total time steps for episode $i$, respectively, then the formula to calculate $reward\_rate$ is defined as:

\begin{equation}
    reward\_rate = \frac{\sum_{i=1}^{N} reward_i} {\sum_{i=1}^{N} time_i}
\end{equation}

\paragraph{\textbf{Average rescuing cost}} represents the total time step cost to earn one unit of reward. This is the inverse of the reward rate.
%\vspace{-4mm}
\begin{equation}
    rescuing\_cost = \frac{1} {reward\_rate}
\end{equation}

% Notice that the model includes the option to have multiple episodes in order to allow us to measure the average performance achieved and the capacity to learn for each rescue policy.
% Algorithm \ref{algorithm:Q-learning} presents the calculation of the total time steps and total rewards per episode.

\subsubsection{Results and Comparisons}
In this work, a heuristic multi-agent reinforcement learning model for disaster relief is trained and evaluated in OpenAI Gym~\cite{1606.01540}. Unlike the standard reinforcement learning settings used for simulations, our experimental environment setting is based on the real-world geographical positions of tweets. Here, a volunteer is formulated as taking action in an environment and receiving rewards and observation at every time step. The training of the agent stops once the policies of volunteers converge. The main purpose is to minimize the amount of time needed to rescue all the victims in the target environment. 

\begin{table}[t]
\setlength{\tabcolsep}{4pt}

  \centering
 \caption{Rescue performance comparison. Bold values represent best performance.}    \begin{tabular}{lr | r | r | r}
    \toprule
          & \multicolumn{1}{ c | }{Time} & \multicolumn{1}{| c |}{Reward} & \multicolumn{1}{| c |}{Reward Rate} & \multicolumn{1}{| c }{Rescuing Cost} \\
  
    \midrule
    Random Walk & 17.4 & 167.2 & 9.6 & 0.104 \\
    \midrule
    Greedy B.F.S & 9.6 & 182.7 & 19.03 & 0.053 \\
    % \midrule
    % A* search & 9.0 & 184.0 & 29.44 & 0.034 \\
    \midrule
    Rule-based & 15.9 & 170.2 & 10.70 & 0.093 \\
    \midrule
    Value Iteration & 14.1 & 173.7 & 12.32 & 0.081 \\
    \midrule
    Reinforcement Learning & 5.4 & 172.0 & 31.85 & 0.031 \\
    \midrule
    Heuristic R. L. & \textbf{5.0} & \textbf{189.7} & \textbf{37.9} & \textbf{0.026} \\
    \bottomrule
    \end{tabular}
  \label{table:performance-comparison}
\end{table}

For these experiments, we transform the geographical distribution of tweets into a grid and set up a centralized communication environment, which consists of N volunteers and M victims in a two-dimensional grid with discrete space and discrete time. The process of extracting geographical information from volunteers and victims is illustrated in Figure~\ref{fig:expriment_geolocation}.
%\textcolor{blue}{The position of volunteers and victims are identified from the tweet feature extraction and then mapped onto the grid} 
Volunteers may take actions in the environment and communicate with the remote central server. 
%There are some random locations hold obstacles due to the disaster that volunteers should avoid going through. 
They will be assigned a penalty if they go off the grid and a reward if they reach the victims they are to rescue.

% {\color{blue}To test our system's performance in real disasters, we set time back to Aug 17, 2017, and start monitor Harvey flooding. At 12:07 AM, early in the morning of Aug 27, 2017, local time, we detect the first tweet requesting help \textit{``I have 2 children with me and tge, water is swallowing us up. Please send help. 111** Sageview Houston, Tx. 911 is not responding!!!!!!"}, which triggers the disaster relief scheduling recommendation function. At this moment, it is in the very first period of reinforcement learning, therefore, the Greedy method can successfully rescue with better total rewards at -100 compared to -450 of the reinforcement learning. However, after the time goes by, a rescue tweet at $Aug 27 xx:yy$, which is $yy$ seconds after the training, the reinforcement rescue can complete the task with higher total rewards at $XXX$ compared to $zzzz$ of the Greedy and $xxxxxx$ of the Random walk method.}

We compared the experimental performance of the proposed \theName{} algorithm with Random walk, Greedy best first search, Rule-based search, Value iteration, and a traditional Reinforcement Learning method.
Figure~\ref{fig:experiment} presents the process of each algorithm's performance within 2000 episode (path from initial to a terminal state). 
In Figure~\ref{fig:expriment_score} and Figure~\ref{fig:expriment_time}, we compare the total rewards and total time steps per episode with each strategy. The \theName{} quickly converges to stable states after the first 24 episodes of training. Once \theName{} converged, it constantly outperforms all other approaches. 
As a comparison, the reinforcement learning technique also performs well after convergence. However, it requires a long time for convergence (208 episodes in current experiment) and the average reward over the entire time period is lower compared to the \theName{}.
The greedy B.F.S strategy performs consistently over the time, shown as points around constant lines. This is not surprising because with this strategy the agents always choose to reach the closest victims first, which is independent of other factors in the rescuing environment. Overall, the reward of greedy B.F.S strategy is less than the \theName{}, while its time steps outperform the \theName{} during the latter's training phase.
The Random walk approach leads to the lowest overall reward as well as the highest completion time per episode, and the performance has a large variation across different episodes. Ruled based and Value iteration are even worse compared to our proposed \theName{} technique.
Figure\ref{fig:expriment_heatmap} and Figure\ref{fig:expriment_box_plot} respectively show the heatmap of the reward distribution and its corresponding box plot. We clearly see that, during the total 2000 testing episodes, the \theName{} has the most of its rewards above 190, while other methods have significantly less number of rewards in this category.

Table~\ref{table:performance-comparison} gives a summary of each algorithm's total time, total reward, reward rate, and rescuing cost. The result clearly shows that the \theName{} has the best overall reward score, the shortest completion time, the highest reward rate, and the lowest rescuing cost rate. In particular, the Greedy B.F.S. and the Reinforced Learning method respectively have the reward and time performance close to the proposed method. Nonetheless, the proposed Heuristic reinforcement learning evidently outperforms these methods when the two metrics are considered simultaneously.

\section{Discussion}
This paper presents a novel algorithm designed to develop a better response to victims' requests for assistance during disasters, along with a case study using Twitter data collected during Hurricane Harvey in 2017. This is one of the first attempts to formulate the large-scale disaster rescue problem as a feasible  heuristic multi-agent reinforcement learning problem using massive social network data. With the proposed method, we can train classifiers to extract victim and volunteer information from tweets and transform the data for use in a reinforcement learning environment. Our key contribution is the design of a heuristic multi-agent reinforcement learning scheduling policy that simultaneously schedules multiple volunteers to rescue disaster victims quickly and effectively. The heuristic multi-agent reinforcement learning algorithm can respond to dynamic requests and achieve an optimal performance over space and time. This approach helps to match volunteers and victims for faster disaster relief and better use of limited public resources. The proposed framework for disaster exploration and relief recommendation is significant in that it provides a new disaster relief channel that can serve as a backup plan when traditional helplines are overloaded.

\bibliographystyle{IEEEtran}
\bibliography{reference} 

\end{document}